\def\paperID{10036}
\def\confName{CVPR}
\def\confYear{2024}

\def\paperTitle{Human-Free Automated Prompting for Vision-Language Anomaly Detection: Prompt Optimization with Meta-guiding Prompt Scheme}
\def\aauthor{
Pi-Wei Chen\\
National Cheng Kung University\\
{\tt\small nf6111015@gs.ncku.edu.tw}

\and
Jerry Lin\\
BakuAI AS\\
{\tt\small jerry@bakuai.no}

\and
Feng-Hao Yeh\\
National Cheng Kung University\\
{\tt\small yeh.feng.hao.110@gmail.com}

\and
Jia Ji \\
National Cheng Kung University\\
{\tt\small jijia20001118@gmail.com}

\and
Zih-Ching Chen \\
NVIDIA\\
{\tt\small virginiac@nvidia.com}

\and
Chao-Chun Chen\\
National Cheng Kung University\\
{\tt\small chencc@imis.ncku.edu.tw}
}

\newif\ifreview 
\newif\ifarxiv \newcommand{\arxiv}{\arxivtrue}
\newif\ifcamera 
\newif\ifrebuttal 

\arxiv 

\pdfoutput=1
\documentclass[10pt,twocolumn,letterpaper]{article}
\ifreview \usepackage[review]{cvpr} \fi
\ifarxiv \usepackage[pagenumbers]{cvpr} \fi
\ifrebuttal \usepackage[rebuttal]{cvpr} \fi
\ifcamera \usepackage{cvpr} \fi

\usepackage{siunitx}
\usepackage{booktabs}
\usepackage{graphicx}	
\usepackage{amsmath}	
\usepackage{amssymb}	
\usepackage{booktabs}
\usepackage{times}
\usepackage{microtype}
\usepackage{epsfig}
\usepackage[table,xcdraw,dvipsnames]{xcolor}
\usepackage{caption}
\usepackage{float}
\usepackage{tabularray}
\usepackage{placeins}
\usepackage{color, colortbl}
\usepackage{stfloats}
\usepackage{enumitem}
\usepackage{tabularx}
\usepackage{xstring}
\usepackage{graphicx} 
\usepackage{multirow}
\usepackage{xspace}
\usepackage{url}
\usepackage{subcaption}
\usepackage{xcolor}
\usepackage[hang,flushmargin]{footmisc}

\ifcamera \usepackage[accsupp]{axessibility} \fi

% ...

\ifarxiv  \fi

\newcommand{\R}[1]{{%
    \textbf{%
        \ifstrequal{#1}{1}{\textcolor{red}{R#1}}{%
        \ifstrequal{#1}{2}{\textcolor{blue}{R#1}}{%
        \ifstrequal{#1}{3}{\textcolor{magenta}{R#1}}{%
        \ifstrequal{#1}{4}{\textcolor{teal}{R#1}}{%
                           \textcolor{cyan}{R#1}%
        }}}}%
    }%
}}

\usepackage{xr-hyper}

\makeatletter
\newcommand*{\addFileDependency}[1]{
  \typeout{(#1)}
  \@addtofilelist{#1}
  \IfFileExists{#1}{}{\typeout{No file #1.}}
}

\makeatother

\definecolor{cvprblue}{rgb}{0.21,0.49,0.74}
\usepackage[pagebackref,breaklinks,colorlinks,citecolor=cvprblue]{hyperref}
\usepackage[capitalize]{cleveref}
\crefname{section}{Sec.}{Secs.}
\crefname{table}{Table}{Tables}
\crefname{figure}{Fig.}{Figs.}

\begin{document}
\title{\paperTitle}
\author{\aauthor}
\maketitle

\begin{abstract}
Pre-trained vision-language models (VLMs) are highly adaptable to various downstream tasks through few-shot learning, making prompt-based anomaly detection a promising approach. Traditional methods depend on human-crafted prompts that require prior knowledge of specific anomaly types. Our goal is to develop a human-free prompt-based anomaly detection framework that optimally learns prompts through data-driven manner, eliminating the need for human intervention.
The primary challenge in this approach is the lack of anomalous samples during the training phase. 
To tackle this challenge, we develop the Object-Attention Anomaly Generation Module (OAGM) to synthesize anomaly samples for training. Furthermore, to prevent learned prompt from overfit on synthesized anomaly feature, we proposed Meta-Guiding Prompt-Tuning Scheme (MPTS) that iteratively adjusts the gradient-based optimization direction of learnable prompts to avoid overfitting to the synthesized anomalies . 
This framework allows for the optimal prompt embeddings by searching in the continuous latent space via backpropagation, free from human semantic constraints. 
Additionally, the modified locality-aware attention improves the precision of pixel-wise anomaly segmentation.
\end{abstract}

\section{Introduction}

In the field of anomaly detection, acquiring annotations for anomalous samples is often challenging. Consequently, prior research has focused on unsupervised learning, which involves training models on large volumes of anomaly-free data to identify deviations from normal features as anomalies. Recently, some approaches have addressed more extreme scenarios where only a few normal samples are available. These methods leverage vision-language pretrained models, such as CLIP, to facilitate few-shot anomaly detection.

\begin{figure}
\setlength{\abovecaptionskip}{0pt}
\setlength{\belowcaptionskip}{0pt}
\centering
\includegraphics[width=\linewidth]{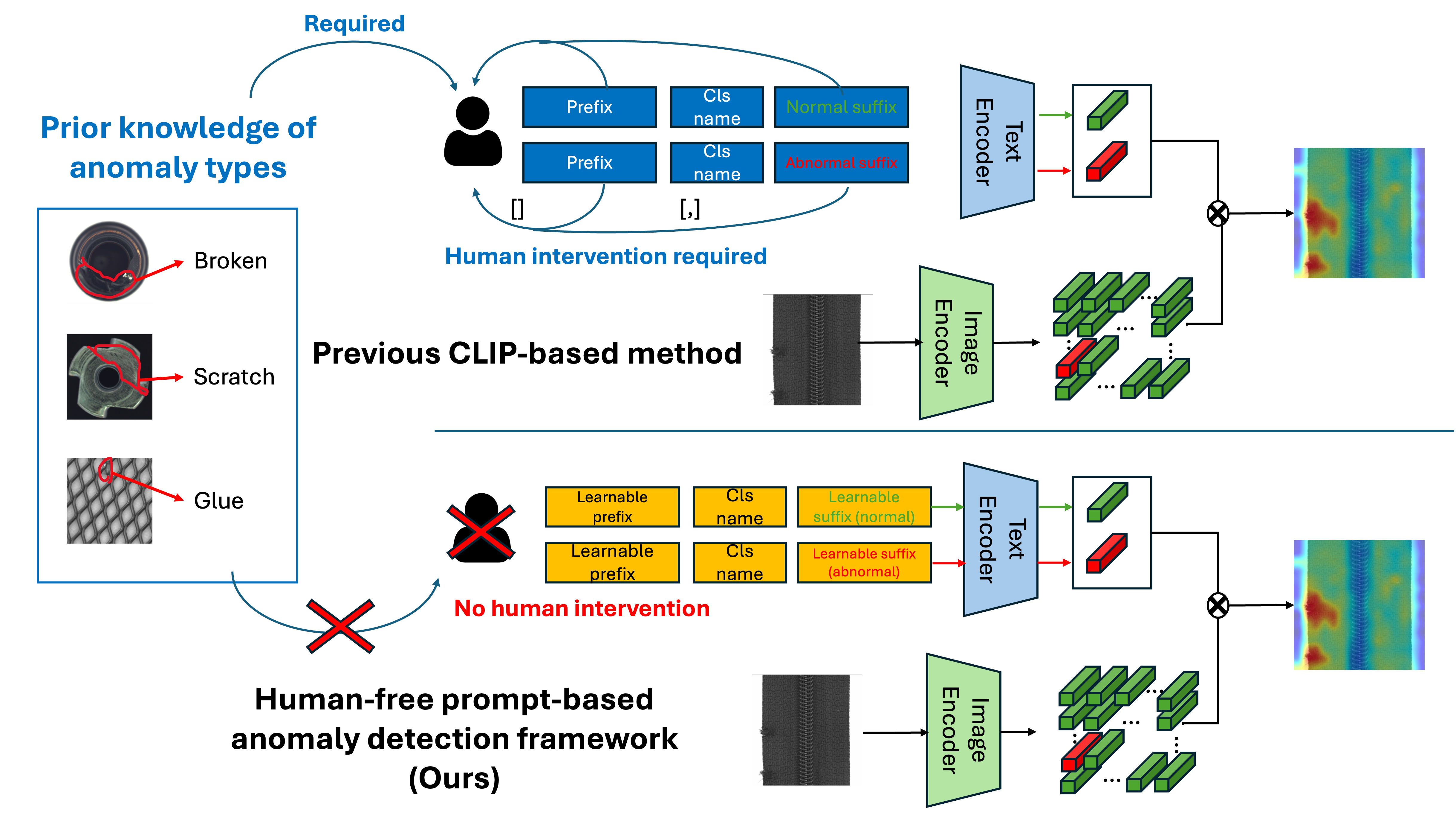}
\caption{The upper part shows previous methods requiring human-designed prompts. The lower part illustrates our automated, data-driven approach using learnable prompts without prior knowledge.}
\label{The illustration of Fig1}
\end{figure}

Vision-language models are renowned for their adaptability to various downstream tasks such as classification and segmentation. The remarkable transferability of vision-language models across different computer vision tasks can be attributed to their large-scale pre-training on text-image pairs \cite{chen2020simple}. This training enables the model to capture relevant semantic features based on the provided prompts, thereby allowing the model to leverage the textual modality to solve visual tasks such as anomaly detection. \cite{jeong2023winclip} pioneered the application of VLMs for zero-shot or few-shot anomaly detection. This approach leverages lexical terms related to 'anomaly' and 'normal' to identify anomalous visual features in the test samples, providing a foundation for subsequent studies to further advance few-shot anomaly detection methodologies.
Prompt engineering is a crucial technique that preserves the robustness and generalization capabilities of CLIP while efficiently adapting it to meet the demands of specific applications. The primary objective of prompt tuning is to obtain an optimal textual embedding that aligns with the target visual image embedding, ensuring that the model accurately captures the nuances of the task at hand. 

Previous work \cite{jeong2023winclip} proposed a prompt ensemble technique that composes different prompt templates (e.g., "a photo of a [c] for visual inspection") and normal-abnormal-relevant prompts (e.g., "flawless" for normality, "damaged" for anomaly) to obtain text embeddings aligned with anomaly detection semantics. Recent studies \cite{li2024promptad, zhou2023anomalyclip} have leveraged prefix prompt tuning (context optimization \cite{zhou2022learning}) to replace static prompt templates with learnable prompts, better catering to anomaly detection tasks. Although prior work has demonstrated the efficacy of prefix prompting, it still necessitates a customized, manually-designed suffix to describe the anomaly type for each class item. This process heavily relies on detailed and specific prior knowledge of the dataset and the types of anomalies that may appear in each product category, as shown in the upper part of Fig~\ref{The illustration of Fig1}. This remains a non-trivial task requiring iterative adjustments, rather than automatically finding the optimal textual embedding.

We are motivated by \cite{zhou2022learning}, which provides insights that the optimal prompts derived through prompt tuning do not necessarily align with human semantics or intuition. Previous work has often focused on combining various human semantic terms, such as 'anomaly' and 'defect,' to generate better embeddings. However, we propose a novel perspective: instead of adhering to human-centric semantics, we should employ data-driven optimization methods, such as gradient-based learning, to derive the most effective textual embeddings. By utilizing these advanced deep learning techniques, we can create more robust and precise embeddings that better capture the nuances needed for anomaly detection and adapt to various datasets without the need for prior knowledge as show in lower part in Fig~\ref{The illustration of Fig1}.

However, to optimize the Learnable Normal Prompt (LNP) and Learnable Anomaly Prompt (LAP), samples from both normal and anomalous data are required. In the case of  few-shot anomaly setting, only normal samples are available during training.
To enable the acquirement of optimal textual embedding for anomaly detection, we need to self-generate anomaly sample to obtain the optimal prompt for better textual embedding.
One intuitive method to synthesize anomalous samples is by adding Gaussian noise to the image pixels to mimic anomalies. However, Gaussian noise alone might not guarantee performance in real-world scenarios because such synthetic anomalies often lack the complex and context-specific characteristics of true anomalies.

Additionally, Vision-Language Models (VLMs) are not designed for pixel-wise tasks like segmentation, making them unsuitable for pixel-level anomaly detection. During VLM pre-training, only global features (e.g., class tokens in Vision Transformers) are considered, neglecting localized features that capture fine-grained image details. Consequently, the relationship between text prompts and specific image regions is not optimized, leading to unpredictable results where the target prompt may not align with the target object pixels \cite{wang2023sclip, li2023clip}.

In this paper, we propose a human-free automated prompting anomaly detection framework, namely \textbf{"Meta-prompting Semantic Learning for Anomaly Detection with Locality-aware Attention Image Encoder"}. The proposed framework follows the intuition of data-driven methods that leverage back-propagation to actively find the prompt that can generate the optimal textual embedding suited for anomaly detection, instead of passively combining all possible human semantic prompts to meet the requirements of prompt-based anomaly detection. To address the limitation of the absence of anomaly samples for context optimization, we develop the \textbf{Object-Attention Anomaly Generation Module} to synthesize anomaly samples for training. Meanwhile, the proposed \textbf{Meta-guiding Prompt-tuning Scheme} (MPTS) is developed to prevent the learnable prompt from overfitting on synthesized anomalies through iterative gradient-based calibration.


In Meta-Prompt Tuning Scheme (MPTS), we introduce a dynamic meta-prompt that serves as an anchor to balance specificity and generality in anomaly detection via gradient-based calibration. This meta-prompt ensures that learnable prompts remain aligned with meaningful, generalizable concepts while leveraging useful information from synthesized anomalies without deviating from the core goal of anomaly detection.
Initially, a general prompt template provided by a large language model (LLM) is embedded with real-world anomaly semantics, serving as the initial meta-prompt. This helps ensure that the learnable prompts do not deviate significantly from true anomalies during fine-tuning.
Throughout each training round, the optimized learnable prompt from the previous round serves as the new meta-guiding prompt, iteratively refining the process. Meta-guiding prompts ensure that the learnable prompts resemble real anomalies, enhancing detection effectiveness. Additionally, introducing Gaussian noise diversifies the patterns, improving generalization across various anomaly types. This combination allows the learnable prompts to generalize well to different anomalies while maintaining alignment with real-world characteristics.

To further improve the quality of synthesized anomalous samples, we introduce the \textbf{"Object-Attention Anomaly Generation Module"}. This module leverages the general knowledge of CLIP to identify the target object in the image and applies noise specifically to that object. By doing so, the synthesized anomalies better reflect practical scenarios, as real-world anomalies usually occur on target objects rather than in the background, which is typically not the focus of anomaly detection.


Additionally, we introduce \textbf{Locality-aware Attention}, which can be adopted by any transformer-based model to mitigate the misalignment of input token features and corresponding output token features. This mechanism restricts attention to local neighborhoods, preserving essential local details and enhancing the alignment between the input features and the output tokens. By focusing on neighboring regions, the locality-aware attention ensures that the feature extraction process maintains the spatial integrity of the image. This addresses the limitations of the vanilla attention mechanism, where distant tokens tend to dilute crucial local information, resulting in the misalignment of locality features between input and output.
The contributions are listed as follows: (1) We propose a novel paradigm for a prompt-based anomaly detection framework, namely \textbf{Meta-prompting Semantic Learning for Anomaly Detection with Locality Feature Attention Image Encoder}, which optimizes the pre-trained VLM for anomaly detection tasks in a human-free prompting manner; (2) We introduce the \textbf{Meta-guiding Prompt-tuning Scheme} along with the \textbf{Object-Attention Anomaly Generation Module} to enable the optimization of learnable prompts in the absence of anomaly samples. The optimized learnable prompts with the \textbf{Meta-guiding Prompt-tuning Scheme} even exceed manually designed prompts by a significant margin; (3) We propose the \textbf{Locality-Aware Transformer}, which aims to extract a feature map where all locality features align with their original image positions, ensuring pixel-wise anomaly segmentation in the CLIP-based framework.


\begin{figure*}

\centering
\includegraphics[width=\linewidth]{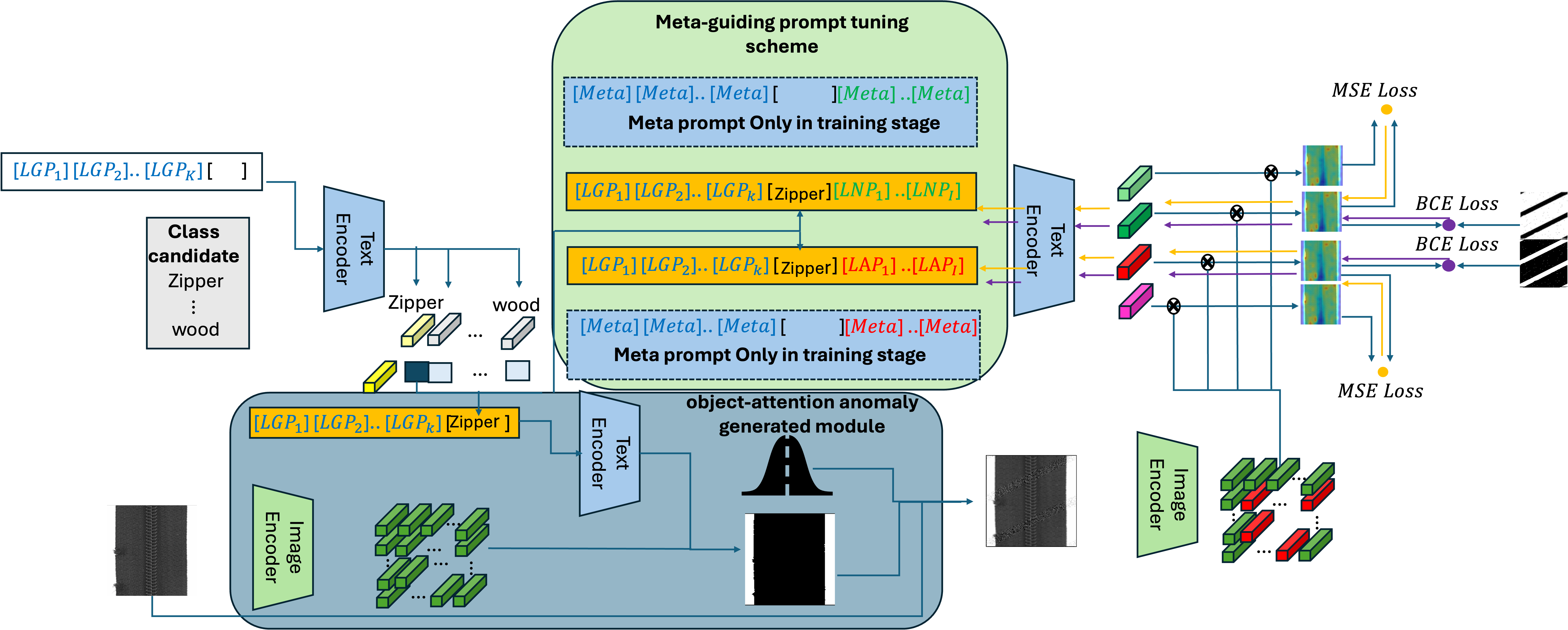}
   \caption{The overall architecture of Meta-prompting Semantic Learning for Anomaly Detection with Locality Feature Attention Image Encoder }
\label{The illustration of overall architecture}
\end{figure*}

\section{Related Work}
\label{sec:related}

\subsection{Vision-language model}

Vision-language models (VLMs) are advanced AI systems that integrate computer vision (CV) and natural language processing (NLP) to understand the commonality of semantics between text embeddings and visual features. These multimodal models are trained on extensive datasets of paired images and text, utilizing techniques like contrastive learning \cite{chen2020simple} and masked language modeling to map visual and textual data into a shared semantic space. A notable open-source VLM is OPENCLIP \cite{ilharco_gabriel_2021_5143773}, which excels in generalization, enabling rapid adaptation to various downstream tasks such as image classification and object detection without task-specific training. Furthermore, \cite{zhou2022learning} proposed context optimization, which can tune learnable prompts with few-shot samples, achieving optimal performance surpassing human-designed prompts with.
\subsection{Anomaly detection}
Traditional paradigms of anomaly detection \cite{xie2016unsupervised, fan2018abnormal, schlegl2019f, chen2022utrad} typically utilize only normal samples for training anomaly detection models. This approach, known as unsupervised anomaly detection, does not require any anomaly samples during training. In recent years, the research focus has shifted from unsupervised anomaly detection to the more challenging few-shot scenario, where only a limited number of normal samples are available during the training phase. PatchCore \cite{cohen2020patchcore} employs a memory-efficient technique to capture normal patterns in few-shot data for anomaly inference. RegAD \cite{zhang2021regad} introduces a region-based method that maximizes the information extracted from each example, effectively learning informative features for anomaly inference using only a few normal samples.

Recent studies \cite{jeong2023winclip, li2024promptad} have further addressed the anomaly detection challenge with multimodal solutions, leveraging the general knowledge embedded in pre-trained vision-language models (VLMs) to perform zero-shot and few-shot anomaly detection with the guidance of textual modality. However, these approaches still rely on human-designed prompts, with their performance dependent on the prompt engineer's prior knowledge of the specific application scenario. This dependency highlights the need for more advanced, data-driven methods to optimize prompt selection and enhance anomaly detection performance.

\section{Proposed Framework}
\label{sec:method}


\subsection{Problem Definition:}
In this paper, we address the few-shot anomaly detection task, where only a few normal samples are available during the training stage. We suppose \( D_{\text{train}} = \{(\mathbf{x}_i, y_i=0)\}_{i=1}^{k} \) is the \( k \)-shot sample given in the training stage (\( y_i \in \{0,1\} \)). Each \( \mathbf{x}_i \in \mathbb{R}^{H \times W} \) is a normal sample when \( y_i=0 \), and abnormal when \( y_i=1 \).

In our proposed human-free prompt-based anomaly detection framework, we substitute traditional human prompts (\textit{"a photo of a \{\} without defect"} as the normality prompt and \textit{"a photo of a defective \{\}"} as the abnormality prompt) with learnable prompts. The learnable normality prompt (LNP) and the learnable abnormality prompt (LAP) are composed of a learnable general prefix (LGP) \( =\left[G_1, \ldots, G_{ng}\right] \) and a learnable normal or abnormal suffix (LNS \( =\left[N_1, \ldots, N_{nr}\right] \) or LAS \( =\left[A_1, \ldots, A_{na}\right] \), respectively:

\begin{equation}
\begin{cases}
\text{LNP} = \left[G_1, \ldots, G_{ng}\right] \left[\text{cls}\right] \left[N_1, \ldots, N_{nr}\right] \ \\
\text{LAP} = \left[G_1, \ldots, G_{ng}\right] \left[\text{cls}\right] \left[A_1, \ldots, A_{na}\right]
\end{cases}
\end{equation}

The LNP and LAP are leveraged to calculate the anomaly score map \( S \in \mathbb{R}^{H \times W \times 1} \) on the feature map \( F \in \mathbb{R}^{H' \times W' \times C} \) extracted from the testing sample.

In the inference stage, we have \( D_{\text{test}} = \{(\mathbf{x}_i, y_i, M_i)\}_{i=1}^{N} \). The mask \( M \in \{0, 1\}^{H \times W} \) is the ground truth map for anomaly pixels, where each pixel \( m_j \in M \) indicates whether the corresponding pixel \( x_j \in x \) is anomalous (\( m_j = 1 \)) or normal (\( m_j = 0 \)).

Our objective is to leverage \( D_{\text{train}} \) to tune LAP and LNP such that they can yield an anomaly score map \( S \) that aligns with the ground truth anomaly map \( M \). This can be represented as:

\begin{equation}
\arg\max \text{Sim}(S, M)
\end{equation}

where \( \text{Sim}() \) is a similarity metric that calculates the alignment degree between \( S \) and \( M \).

\subsection{Overview}
The overall framework of the proposed \textbf{Meta-prompting Semantic Learning for Anomaly Detection with Locality Feature Attention Image Encoder} is illustrated in Fig~\ref{The illustration of overall architecture}. The proposed Meta-guiding Prompt-tuning Scheme (Sec~\ref{sec:Meta-guiding Prompt-tuning Scheme}) sets a new paradigm for prompt-based anomaly detection with no human semantic lexicon involved in inference. 
To address the absence of anomalous samples for tuning prompts, the proposed Object-Attention Anomaly Generation Module (OAGM) (Sec~\ref{sec:Object-attention Anomaly Generation Module}) produces synthesized anomalous samples for tuning while remaining practical by only applying to items instead of the background. 
Additionally, to enhance the locality feature extraction, the visual encoder adopts a novel proposed Locality-Aware Transformer (Sec~\ref{sec:Locality-aware attention}), which enhances locality feature extraction with neighbor-cls-only attention. 
The prompts  LG and normal/abnormal suffix ( LNS / LAS ) are designed as fully learnable parameters.
\subsection{Meta-guiding Prompt-tuning Scheme}
\label{sec:Meta-guiding Prompt-tuning Scheme}
\begin{figure}
\setlength{\abovecaptionskip}{0pt}
\setlength{\belowcaptionskip}{0pt} 
\centering
\includegraphics[width=\linewidth]{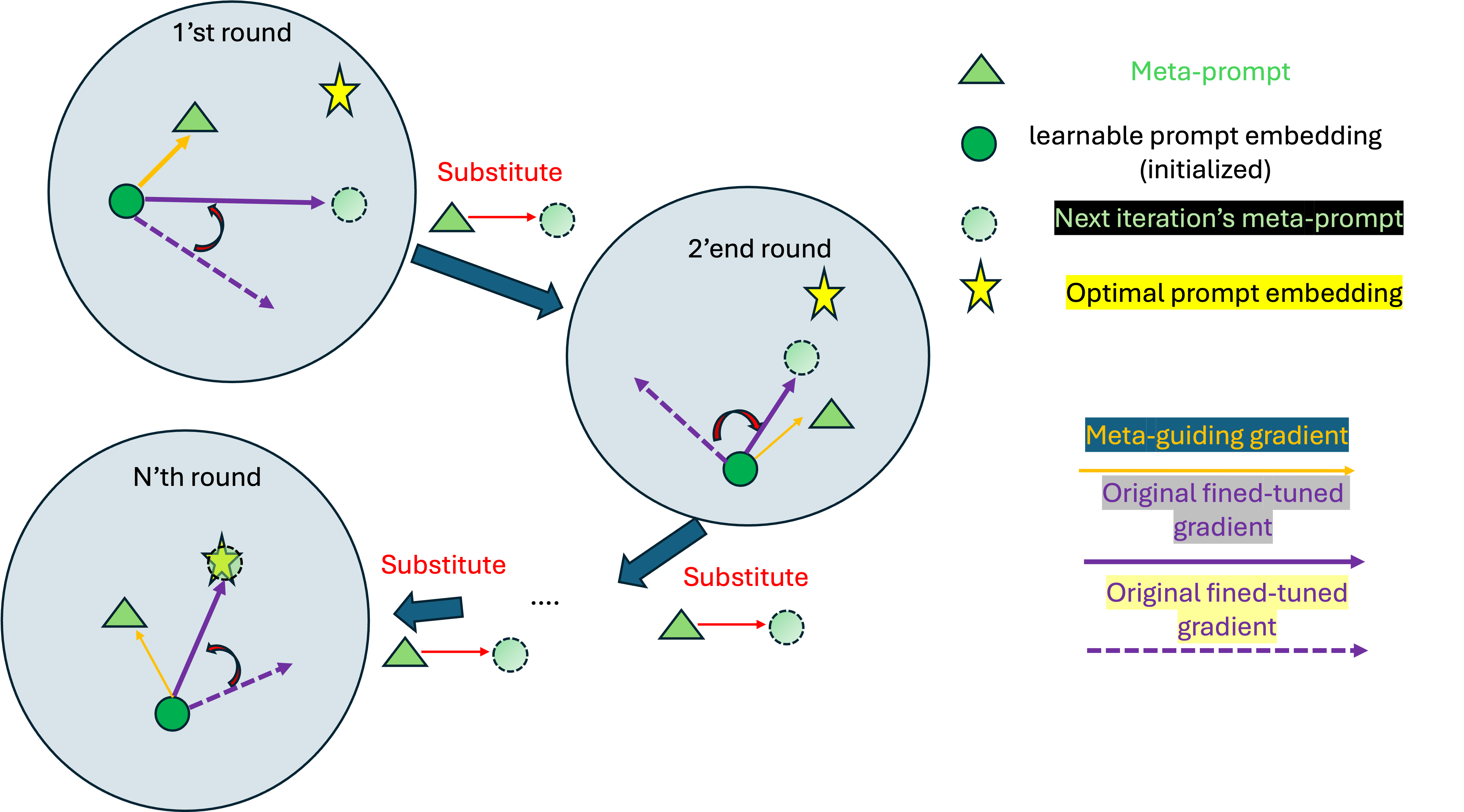}
   \caption{The illustration of Meta-guiding Prompt-tuning Scheme}
\label{The illustration of Meta-guiding Prompt-tuning Scheme}
\end{figure}
The Meta-guiding Prompt-tuning Scheme is illustrated in Fig~\ref{The illustration of Meta-guiding Prompt-tuning Scheme}, which an self-optimized meta-prompt  will iterativly update it self and guiding the learning direction of learnable prompt with gradient correction.

To tune LAP and LNP to align with the anomaly detection task, the OAGM (Sec~\ref{sec:Object-attention Anomaly Generation Module}) will synthesize anomalous samples ${(\mathbf{x'}_i, y_i=1, M'a)}{i=1}^{k}$, where $M'_a \in {0, 1}^{H \times W}$ is the abnormality binary mask indicating whether each pixel $m_j \in M'_a$ is contaminated with synthesized anomaly ($m_j = 1$) or not ($m_j = 0$). A normality binary mask $M'_n$ can be obtained by:
\begin{equation}
M'_n = 1 - M'_a
\end{equation}
The LNP with the extracted feature map of \( \mathbf{x'}_i \) is dot-multiplied to calculate the normality score map \( S_n \), indicating whether each pixel in \( \mathbf{x'}_i \) aligns with normality. Similarly, LAP is used to obtain an abnormality score map \( S_a \).
To aggregate the information from both LNP and LAP and produce a probability-like output (bounded from 0 to 1), we concatenate \( S_n \) and \( S_a \) and apply Softmax along each element, finally obtaining normalized score maps \( S'_n \) and \( S'_a \).
The training objective is to tune LNP such that its embedding perfectly aligns with the normality embedding in \( \mathbf{x'} \). This involves aligning \( S'_n \) with the normality binary mask \( M'_n \) using BCEloss:
\begin{align}
\label{LNP_tuning}
\mathcal{L}_{\text{ano}}^{\text{LNP}} = -  \sum_{j \mid m_j = 1} \Big[ & M'_n(j) \log S'_n(j) \notag \\
& + (1 - M'_n(j)) \log (1 - S'_n(j)) \Big]
\end{align}
For LAP, the objective is to align LAP with the abnormality embedding in \( \mathbf{x'} \):
\begin{align}
\label{LAP_tuning}
\mathcal{L}_{\text{ano}}^{\text{LAP}} = -  \sum_{j \mid m_j = 1} \Big[ & M'_a(j) \log S'_a(j) \notag \\
& + (1 - M'_a(j)) \log (1 - S'_a(j)) \Big]
\end{align}

\textbf{Meta-guiding}:
Synthesized anomaly samples, while useful for training, may contain artifacts or specific characteristics that are not representative of real-world anomalies. To address this limitation, we propose a meta-guiding scheme to regularize the tuning process with general, domain-agnostic knowledge embedded in pre-trained VLMs.

Initially, we provide general manual prompts as Meta Normality Prompt (MNP) and Meta Abnormality Prompt (MAP) (see appendix), which act as anchors to prevent LAP and LNP from overfitting on artifact anomalies and maintain generalization with a broader concept of anomaly. By calibrating the LNP and LAP gradients from Eq~\ref{LNP_tuning} and Eq~\ref{LAP_tuning}, respectively, with the divergence loss between MAP/MNP and LAP/LNP, we can ensure that LAP and LNP are optimized toward a task-specific prompt that human semantics cannot achieve while maintaining generalization.

This divergence loss is designed to estimate the output score maps from LAP/LNP (\( S'_a \) and \( S'_n \), respectively) and MAP/MNP (denoted as \( S^{M}_a \) and \( S^{M}_n \), respectively). Since the values of the output score maps are already transformed to probability form, we use KL divergence as the distance metric to estimate the divergence. The divergence loss can be written as:
\begin{equation}
\mathcal{L}_{\text{div}}^{A} = \sum_{i=1}^{H \times W} \text{KL}(S^{M}_a(i) \parallel S'_a(i))
\end{equation}

\begin{equation}
\mathcal{L}_{\text{div}}^{M} = \sum_{i=1}^{H \times W} \text{KL}(S^{M}_n(i) \parallel S'_n(i))
\end{equation}

To strike a balance between task-specific objectives and generalization, the meta-guiding scheme will only calibrate the gradient when the gradient \( G_{\text{ano}}^{\text{LAP/LNP}} \) from fine-tuning the anomaly loss:
\begin{equation}
    G_{\text{ano}}^{\text{LAP/LNP}} = \frac{\partial \mathcal{L}^{\text{LAP/LNP}}_{\text{ano}}}{\partial \theta_{\text{LAP/LNP}}}
\end{equation}
contradicts the gradient from divergence loss \( G_{\text{div}}^{\text{LAP/LNP}} \).
\begin{equation}
    G_{\text{div}}^{\text{LAP/LNP}} = \frac{\partial \mathcal{L}_{\text{div}}^{\text{A/M}}}{\partial \theta_{\text{LAP/LNP}}}
\end{equation}
The LAP and LNP are optimized with the gradient \( G_{\text{ano}}^{\text{LAP/LNP}} \) after being calibrated with Eq~\ref{eq:guiding}:
\begin{equation}
\label{eq:guiding}
    \begin{cases} 
        G_{\text{ano}} - \lambda \cdot \text{Cos}\left(G_{\text{ano}}, G_{\text{div}}\right), & \text{if } \text{Cos}\left(G_{\text{ano}}, G_{\text{div}}\right) < 0 \\
        G_{\text{ano}}, & \text{otherwise}
    \end{cases}
\end{equation}
The meta-guiding scheme ensures that the optimization direction will explore the optimal prompt that aligns better with the anomaly detection task while preventing overfitting on synthesized anomalies.

Based on the intuition that we have obtained a more optimal \( \text{LAP}^{t+1} \)/\( \text{LNP}^{t+1} \) through gradient descent instead of relying on human semantics, we substitute the MAP/MNP with \( \text{LAP}^{t+1} \)/\( \text{LNP}^{t+1} \) as the new Meta prompt for the next meta-epoch training. We iteratively refine the MAP/MNP and LAP/LNP until the loss converges.
This iterative substitution strategy balances the need for task-specific adaptation and generalization. By continually referring back to the meta-guiding prompts, the model maintains a connection to the initial general knowledge while fine-tuning the prompts to better detect anomalies.

\subsection{Object-attention Anomaly Generation Module}
\label{sec:Object-attention Anomaly Generation Module}
To tune LAP and LNP to align with the anomaly detection task, we propose the Object-Attention Anomaly Generation Module (OAGM). Traditional methods that add Gaussian noise randomly across the entire image are suboptimal because they affect irrelevant background regions, diluting the model's ability to focus on the target object. The OAGM addresses this limitation by selectively adding noise only to the target object areas, enhancing the model's ability to detect anomalies.

First, we utilize a pre-trained Vision-Language Model (VLM) to produce the binary mask $\mathbf{M}_\text{obj}$ for the target object:
\[
\mathbf{M}_\text{obj}(i,j) = 
\begin{cases} 
1 & \text{if pixel } (i,j) \text{ belongs to the target object} \\
0 & \text{otherwise}
\end{cases}
\]

Next, Gaussian noise $\mathcal{N}(0, \sigma^2)$ is selectively added to the pixels identified by $\mathbf{M}_\text{obj}$:
\[
\mathbf{x}'(i,j) = 
\begin{cases} 
\mathbf{x}(i,j) + \mathcal{N}(0, \sigma^2) & \text{if } \mathbf{M}_\text{obj}(i,j) = 1 \\
\mathbf{x}(i,j) & \text{if } \mathbf{M}_\text{obj}(i,j) = 0
\end{cases}
\]

\subsection{Locality-aware attention}
\label{sec:Locality-aware attention}
The vanilla ViT is not suitable for pixel-level tasks due to the \textbf{feature misalignment} issue. We obtain insight from \cite{dosovitskiy2020image} that the feature misalignment issue results from the nature of the original attention mechanism, where each patch (token) feature attends to distant patch features. This leads to local features being fused with irrelevant background features, causing misalignment between input and output features.

Based on this observation, we substitute the vanilla attention with the proposed Locality-Aware Attention (LAA) that restricts each token to attend only to its neighboring token features during feature extraction. To avoid model collapse caused by directly modifying the self-attention mechanism, which would lead to changes in subsequent layers and be amplified through the layers, we adopt the dual-path design from \cite{} to maintain stable input features by preserving the original path alongside the new path.

The LAA applies a $k$-neighbor exclusive mask $\mathbf{M}_{\text{LAA}} \in \mathbb{R}^{H' \times W'}$ to mask the attention map (the dot product of $\mathbf{Q}$ and $\mathbf{K}$) in the attention mechanism:
\begin{equation}
    \text{Attention}_{\text{LAA}}(\mathbf{Q}, \mathbf{K}, \mathbf{V}) = \text{softmax} \left( \frac{\mathbf{Q} \mathbf{K}^T + \mathbf{M}_{\text{LAA}}}{\sqrt{d_k}} \right) \mathbf{V}
\end{equation}
where $\mathbf{M}_{\text{LAA}}$ is constructed with the following formula to enable each token to attend only to its $k$-neighbors:
\begin{equation}
    M_{(i,j),(m,n)} =
\begin{cases}
0 & \text{if } \sqrt{(i - m)^2 + (j - n)^2} \leq k \\
-\infty & \text{otherwise}
\end{cases}
\end{equation}
Each patch can be indexed in the 2D grid as $(i, j)$ where $0 \leq i < H'$ and $0 \leq j < W'$.
$(i, j)$ and $(m, n)$ are the coordinates of the patches in the original feature map before patchification, and $k$ is the predefined neighbor distance.

\section{Experiments}
\label{sec:experiment}
We conduct experiment to validate that Meta-guiding Prompt-tuning Scheme with Object-attention Generation Module can optimize a prompt embedding that outperform human-semantic prompt under scenario that only 1 sample (1-shot) is provided in pixel-level anomaly segmentation. 
Additionally, we complete ablation  experiment to verify the effectiveness of each proposed component. \\
\textbf{Dataset}
In this paper, we conduct experiment on MVTec\cite{bergmann2019mvtec} dataset, which is a benchmark dataset for industrial anomaly detection that provide pixel-level ground truth map indicating the anomaly location in the image. MVtec comprise 15 Categories of item, and each category contain 5-6 sub-categories of anomlay type.
\subsection{Comparison to human-semantic prompt}
\begin{table*}
\centering
\caption{Comparison to Human semantic prompt}
\label{tab:Comparison to Human semantic prompt}
\resizebox{\textwidth}{!}{
\begin{tabular}{lrrrrrrrrrrrrrrrr}
\toprule
Class &  carpet &   grid &  leather &   tile &   wood &  bottle &  cable &  capsule &  hazelnut &  metal\_nut &   pill &  screw &  toothbrush &  transistor &  zipper &   mean \\
\midrule
Human-semantic prompt &   93.98 &  35.86 &    90.89 &  43.21 &  57.74 &   82.86 &  42.54 &    72.05 &     84.09 &      55.98 &  51.57 &  70.77 &       75.43 &       62.23 &   50.90 &  64.67 \\
Ours    &   99.40 &  98.45 &    99.38 &  96.12 &  96.18 &   89.91 &  87.75 &    94.51 &     97.70 &      75.08 &  95.92 &  95.48 &       97.23 &       72.23 &   90.63 &  92.40 \\
\bottomrule
\end{tabular}}
\end{table*}
As shown in Table\ref{tab:Comparison to Human semantic prompt}, we can see that our proposed framework outperform manual design prompt by 28\%, which further validate our assumption that optimal prompt for downstream task is not necessary align with human semantic.
The human-semantic prompt template and suffix is followed by the literature \cite{}.
We have further conduct ablation study in Sec.~\ref{sec:Ablation Study} to verify that the meta-prompt guiding scheme is crucial for learning the optimal prompt embedding.

\subsection{Ablation Study}
\label{sec:Ablation Study}
\textbf{Ablation on Meta-prompt guiding scheme}
\begin{table}[h]
\centering
\caption{The ablation study of guiding level $\lambda$ for Meta-guiding Prompt-tuning Scheme}
\label{tab:meta_guiding_ablation_study}
\begin{tabular}{l|c|c|c}
\hline
\multicolumn{4}{c}{Ablation study on MPTS} \\
\hline
Class &  $\lambda=0$ &  $\lambda=0.5$ & $\lambda=1$ \\
\hline
carpet & 76.27 & 99.32 & 99.40 \\
grid & 60.51 & 97.97 & 98.45 \\
leather & 76.84 & 99.15 & 99.38 \\
tile & 80.87 & 94.86 & 96.12 \\
wood & 50.06 & 95.42 & 96.18 \\
bottle & 49.01 & 84.26 & 89.91 \\
cable & 48.90 & 81.80 & 87.75 \\
capsule & 87.55 & 91.84 & 94.51 \\
hazelnut & 77.84 & 87.74 & 97.70 \\
metal\_nut & 52.74 & 76.77 & 75.08 \\
pill & 67.36 & 87.33 & 95.92 \\
screw & 78.54 & 95.27 & 95.48 \\
toothbrush & 52.90 & 92.59 & 97.23 \\
transistor & 60.55 & 65.23 & 72.23 \\
zipper & 88.22 & 93.90 & 90.63 \\
\hline
mean & 67.21 & 89.56 & 92.40 \\
\hline
\end{tabular}
\end{table}

The ablation study results in Table\ref{tab:meta_guiding_ablation_study} highlight the effectiveness of the Meta-guiding Prompt-tuning Scheme, particularly in enhancing anomaly detection performance. When $\lambda=0$, the model lacks meta-guiding and achieves a mean score of 67.21, indicating limited effectiveness due to overfitting on synthesized anomalies. Introducing meta-guiding with $\lambda=0.5$ significantly improves the mean score to 89.56, demonstrating enhanced generalization and alignment with broader anomaly concepts. The highest performance is achieved with $\lambda=1$, yielding a mean score of 92.40 and near-perfect results in several classes, confirming that the scheme effectively balances task-specific tuning and general knowledge integration. These findings validate the module's contribution to improving anomaly detection accuracy and robustness.\\
\textbf{Ablation on Object-attention Anomaly Generated Module(OAGM)}

\begin{table}[h]
\centering
\caption{Comparison of Anomaly Detection Performance with and without OAGM}
\label{tab:oagm_comparison}
\begin{tabular}{l|c|c}
\hline
\multicolumn{3}{c}{Ablation study on OAGM} \\
\hline
Class & Without OAGM & With OAGM ($\lambda=1$) \\
\hline
carpet & 98.36 & 99.40 \\
grid & 96.53 & 98.45 \\
leather & 97.93 & 99.38 \\
tile & 91.14 & 96.12 \\
wood & 89.19 & 96.18 \\
bottle & 64.18 & 89.91 \\
cable & 49.31 & 87.75 \\
capsule & 85.53 & 94.51 \\
hazelnut & 65.62 & 97.70 \\
metal\_nut & 72.12 & 75.08 \\
pill & 78.42 & 95.92 \\
screw & 87.06 & 95.48 \\
toothbrush & 77.05 & 97.23 \\
transistor & 66.85 & 72.23 \\
zipper & 75.61 & 90.63 \\
\hline
mean & 78.54 & 92.40 \\
\hline
\end{tabular}
\end{table}
The ablation study in Table\ref{tab:oagm_comparison} demonstrates the effectiveness of the Object-attention Anomaly Generation Module (OAGM) across different categories. For texture items like carpet, wood, and tile, the entire image serves as the target, leading to high baseline performance even without OAGM (98.36, 89.19, and 91.14, respectively). OAGM provides slight improvements (99.40, 96.18, and 96.12) due to the inherent nature of these items. Conversely, object-like items such as bottle, cable, and capsule show significant performance boosts with OAGM, increasing from 64.18, 49.31, and 85.53 to 89.91, 87.75, and 94.51, respectively. The overall mean score rises from 78.54 to 92.40, highlighting OAGM's role in enhancing anomaly detection by focusing on target objects and reducing irrelevant background noise, especially in object-centric categories.\\
\textbf{Ablation on Locality-aware Attention}\\
To evaluate the impact of the proposed Locality-Aware Attention (LAA), we conducted an ablation study comparing different configurations of attention mechanisms. Specifically, we compared the original VV-att (V-V attention as described in \cite{li2023clip}), QK-LAA (original QK attention with LAA), and VV-LAA (V-V attention combined with LAA). The results are summarized in Table\ref{tab:Ablation study on Locality-Aware Attention}.

\begin{table}[h]
\centering
\caption{Ablation study on Locality-Aware Attention}
\label{tab:Ablation study on Locality-Aware Attention}
\begin{tabular}{l|c|c|c}
\hline
\multicolumn{4}{c}{Ablation study on Locality-Aware Attention} \\
\hline
Class & VV-att & QK-LAA & VV-LAA \\
\hline
carpet & 98.29 & 99.29 & 99.40 \\
grid & 95.93 & 96.93 & 98.45 \\
leather & 98.10 & 99.10 & 99.38 \\
tile & 94.63 & 95.63 & 96.12 \\
wood & 94.07 & 95.07 & 96.18 \\
bottle & 79.83 & 80.83 & 89.91 \\
cable & 84.58 & 85.58 & 87.75 \\
capsule & 89.09 & 90.09 & 94.51 \\
hazelnut & 94.53 & 95.53 & 97.70 \\
metal nut & 72.64 & 73.64 & 75.08 \\
pill & 85.12 & 86.12 & 95.92 \\
screw & 91.77 & 92.77 & 95.48 \\
toothbrush & 95.55 & 96.55 & 97.23 \\
transistor & 70.19 & 71.19 & 72.23 \\
zipper & 87.45 & 88.45 & 90.63 \\
\hline
mean & 89.52 & 90.52 & 92.40 \\
\hline
\end{tabular}
\end{table}

From the results, it is evident that LAA enhances performance across all categories. The mean performance of QK-LAA (90.52) surpasses that of VV-att (89.52), demonstrating that the locality constraint improves feature alignment even when applied to the traditional QK attention mechanism. Furthermore, the combination of V-V attention with LAA (VV-LAA) achieves the highest mean performance (92.40), indicating that our LAA not only addresses the feature misalignment issue but also synergizes effectively with advanced attention mechanisms to boost overall performance. This significant improvement underscores the effectiveness of LAA in enhancing pixel-wise anomaly detection tasks.
\section{Conclusion}
\label{sec:conclusion}
In this paper, we introduced a novel framework, \textbf{Meta-prompting Semantic Learning for Anomaly Detection with Locality Feature Attention Image Encoder}, which optimizes vision-language models for anomaly detection in a human-free manner. Our \textbf{Meta-guiding Prompt-tuning Scheme} and \textbf{Object-Attention Anomaly Generation Module} address the absence of anomaly samples by synthesizing realistic anomalies and iteratively refining prompts via gradient-based calibration. Additionally, the \textbf{Locality-Aware Transformer} preserves local details and ensures alignment between input features and output tokens, enhancing pixel-wise anomaly segmentation. Our extensive experiments demonstrated significant performance improvements over manually designed prompts, showcasing the efficacy of our approach and advancing the field of few-shot and zero-shot anomaly detection.This work opens new avenues for future research in developing more sophisticated anomaly detection models that can adapt to various real-world scenarios without extensive human intervention.
\label{sec:conclusion}


{\small
\bibliographystyle{ieeenat_fullname}
\bibliography{11_references}
}

\end{document}


\title{\paperTitle}
\author{\authorBlock}
\maketitlesupplementary

\textbf{Appendix Section}
\section{Complexity Analysis of Image Compositions}
\begin{table}[h]
\centering
\begin{tabular}{|l|l|l|l|}
\hline
            Dataset type        & MVTec & CIFAR10 & Mnist \\ \hline
Image Entropy value &    5.6   & 4.5     & 1.32  \\ \hline
\end{tabular}
\caption{The comparison table of the mean image entropy value for MVTec, CIFAR10, Mnist}
\label{tab:image_entropy}
\end{table}

In Sec. \ref{sec:topk}, we state that the complexity of the image compositions within the datasets significantly influences the optimal choice for \(k\).
Since the image entropy can be used to quantify and measure the information content, pixel distribution, randomness and structure of the image, we use it to estimate the complexity of the image compositions within the datasets.
It is calculated using the following steps:

\begin{enumerate}
  \item We calculate the histogram of the pixel values in the image. For a grayscale image with pixel values between 0 and 255, the histogram \( h \) has 256 entries, where each entry \( h(i) \) represents the number
 of pixels with the intensity \( i \).
  \item We normalize the histogram to obtain the probability distribution \( p \) of the intensity levels. To do this, each histogram number \( h(i) \) is divided by the total number of pixels \( N \) in the image:
  \[
  p(i) = \frac{h(i)}{N}
  \]
  where \( N = \sum_{i=0}^{255} h(i) \).
  \item Calculate the entropy \( H \) using the probability distribution:
  \[
  H = -\sum_{i=0}^{255} p(i) \log_2 p(i)
  \]
  where the sum is taken over all possible intensity levels, and \( \log_2 \) denotes the base-2 logarithm.
\end{enumerate}

Entropy \( H \) is a non-negative value that quantifies the average amount of information or uncertainty per pixel in the image. A higher entropy value indicates a more complex image with a greater variety of pixel values, while a lower entropy value suggests a simpler image with less variability in pixel values.

{\color{red}Table \ref{tab:image_entropy}} shows that the image entropy of MVTec and CIFAR10 is relatively higher than that of Mnist, which indicates that the image composition of Mnist is less complex than that of MVTec and CIFAR10.

\section{Statistical Modeling of Latent Space Boundaries in Relation to Dataset Variability}
\label{app:b}
\begin{table*}[]
\centering
\caption{The probability density value of the anomalous embedding from the multi-class PDF is on average higher than the single-class PDF, indicating that the model trained on multi-classes tends to generalize covariate anomalies.}
\resizebox{\textwidth}{!}{%
\begin{tabular}{|c|c|c|c|c|c|c|c|c|c|c|c|c|c|c|c|}
\hline
\multicolumn{16}{|c|}{Probability density value on MVTec} \\
\hline
& Tile & Leather & Bottle & Grid & Transistor & Wood & Screw & Hazelnut & Cable & Zipper & Metal nuts & Pill & Capsule & Carpet & \textbf{Mean} \\
\hline
Single-class & 0 & 0 & 0 & 0 & 0 & 0 & 0 & 0 & 0 & 0 & 0 & 0 & 0 & 0 & \textbf{0} \\
\hline
Multi-class & 3.1 & 1.0 & 1.4 & 4.3 & 2.3 & 4.5 & 2.5 & 5.4 & 9.6 & 5.9 & 8.0 & 1.0 & 5.8 & 3.3 & \textbf{4.1} \\
\hline
\end{tabular}%
}
\label{tab:MVTec_PDF_comparison}
\end{table*}

\begin{table*}[]
\setlength{\abovecaptionskip}{0pt}
\setlength{\belowcaptionskip}{0pt}
\caption{The probability density value of the anomalous embedding from the multi-class PDF is on average higher than the single-class PDF, suggesting that a multi-class based model tends to have a more general boundary for anomalous sampling in the latent space, especially for semantic anomalies.}
\begin{center}
\begin{tabular}{|c|cc|cc|cc|}
\hline
\multicolumn{7}{|c|}{\textbf{Probability density value on MVTec}} \\ \hline
\multicolumn{1}{|c|}{Class} & \multicolumn{2}{c|}{\{0,1,2,3,4\}} & \multicolumn{2}{c|}{\{1,3,5,7,9\}} & \multicolumn{2}{c|}{\{0,2,4,6,8\}} \\ \cline{2-7} 
& Mnist & CIFAR10 & Mnist & CIFAR10 & Mnist & CIFAR10 \\ \hline
Single-class& 0 & 0 & 0 & 0 & 0 & 0 \\ \hline
Multi-class & 2.3 & 2.1 & 2.5 & 3.2 & 2.8 & 2.9 \\ \hline
\end{tabular}
\label{tab:Mnist_CIFAR10_comparison}
\end{center}
\end{table*}

Each dataset has a unique statistical distribution that captures the variations and characteristics of its features. A reconstruction model is trained to capture this dataset and encode it in a latent space. This encoded latent representation is characterized by its own statistical properties, including a mean and variance that correspond to the distribution of the original dataset.
If the embeddings fall within this distribution in the latent space, they are considered to lie within a defined boundary — the region where the model expects the projections of the training data to lie.

As hypothesis shown in Sec. 3.2, the models trained on multi-class datasets exhibit more diverse latent representations due to the greater variability of the training data. This diversity broadens the distribution of the latent space, so that a wider range of embeddings may be considered normal. As a result, the likelihood that an anomalous embedding is within the boundary of normality of the latent space increases, potentially leading to lower sensitivity in detecting anomalies.

To substantiate this claim, we systematically conduct empirical experiments to substantiate this claim.
The experiment involves modeling the data embeddings of both single-class and multi-class settings from the datasets using a Gaussian mixture model (GMM). This allows us to estimate the Probability Density Functions (PDFs) that describe the likelihood that an anomalous embedding falls within the normal embedding distribution in the latent space:
\begin{itemize}
    \item \textbf{High probability density values:} The higher values of probability density for the anomaly sample indicate that anomalous embeddings are more likely to fall in regions of high PDF probability. This suggests that the model's valid boundary for the normal data embedding was shaped to encompass a wide range of variations, including those characteristic of anomalies.
    \item \textbf{Low probability density values:} Lower probability density values for an anomaly sample would indicate that the anomalous embeddings are more likely to fall in regions of low PDF probability. This suggests that the latent space of the model has a boundary that does not encompass a wide range of variation and is instead more closely tied to the characteristics of the normal data.
\end{itemize}

\subsection{Experimental setting}
The experiments were performed with the MVTec, CIFAR10 and MNIST datasets. For the single-class setting, the models were trained on data from one class only and the resulting embeddings were used to construct a Gaussian Mixture Model (GMM). The multi-class setting follows the procedures described in Sec. \ref{sec:experiment}.

To investigate whether training on a multi-class dataset increases the likelihood that covariate anomalies are captured by the valid latent space, we used embeddings of defective elements of a given class in the MVTec dataset as input to the GMM-derived probability density functions (PDFs). In terms of semantic anomalies, embeddings of classes that are considered anomalous (for example, if classes {0,1,2,3,4} are considered normal, then classes {5,6,7,8,9} are considered anomalous) were similarly tested against the estimated PDFs.

\subsection{Experiment result}
According to Table \ref{tab:MVTec_PDF_comparison}, the PDFs associated with the multi-class configuration for the MVTec dataset show a 4.9\% increase in density values attributed to anomalies compared to single-class models, indicating a more variant valid boundary in latnet space.

Moreover, Table \ref{tab:Mnist_CIFAR10_comparison} reflects that, in the Mnist and CIFAR10 datasets, single-class PDFs virtually eliminate anomalies, whereas multi-class PDFs assign significantly higher density values to these outliers, signaling an extension in the boundary's inclusivity.

\subsection{Conclusion}
Based on the above experimental result, we can statistically confirm the assertion made in Sec.\ref{sec:Discussion of the behavior of reconstruction network in multi-class dataset} that a reconstruction model trained on multiclass datasets inevitably leads to a more variant valid boundary that includes both the covariate and the semantic embedding.
\begin{table*}[]
\setlength{\abovecaptionskip}{0pt}
\setlength{\belowcaptionskip}{0pt} 
\caption{Evaluation of AUROC for three different settings (\{0,1,2,3,4\}, \{1,3,5,7,9\}, \{0,2,4,6,8\}) on CIFAR10 and Mnist}
\begin{center}
\begin{tabular}{|ccccccccccc|}
\hline
\multicolumn{11}{|c|}{\textbf{AUROC performence on Mnist and CIFAR10}}                                                                                                                    \\ \hline
\multicolumn{1}{|c|}{\multirow{2}{*}{}} & \multicolumn{2}{c|}{US \cite{bergmann2020uninformed}}   & \multicolumn{2}{c|}{FCDD+OE \cite{liznerski2020explainable}} & \multicolumn{2}{c|}{PANDA\cite{mishra2021vt}}          & \multicolumn{2}{c|}{MKD \cite{salehi2021multiresolution}}           & \multicolumn{2}{c|}{Ours} \\ \cline{2-11} 
\multicolumn{1}{|c|}{}                  & Mnist & \multicolumn{1}{c|}{CIFAR10}& Mnist & \multicolumn{1}{c|}{CIFAR10}& Mnist & \multicolumn{1}{c|}{CIFAR10} & Mnist & \multicolumn{1}{c|}{CIFAR10} & Mnist      & CIFAR10      \\ \hline
\multicolumn{1}{|c|}{\{0,1,2,3,4\}}     & 0.57 & \multicolumn{1}{c|}{0.54}& 0.65 & \multicolumn{1}{c|}{0.75} & 0.64 & \multicolumn{1}{c|}{0.68} & 0.69  & \multicolumn{1}{c|}{0.78} & \textbf{0.99} & \textbf{0.94} \\ \hline
\multicolumn{1}{|c|}{\{1,3,5,7,9\}}    & 0.55 & \multicolumn{1}{c|}{0.57}& 0.63 & \multicolumn{1}{c|}{0.79} & 0.59 & \multicolumn{1}{c|}{0.70} & 0.59 & \multicolumn{1}{c|}{0.70} & \textbf{0.86} & \textbf{0.91} \\ \hline
\multicolumn{1}{|c|}{\{0,2,4,6,8\}}    & 0.53 & \multicolumn{1}{c|}{0.54}& 0.64 & \multicolumn{1}{c|}{0.74} & 0.63 & \multicolumn{1}{c|}{0.78} & 0.55 & \multicolumn{1}{c|}{0.68} & \textbf{0.85} & \textbf{0.98} \\ \hline
\multicolumn{1}{|c|}{mean}               & 0.55 & \multicolumn{1}{c|}{0.55}& 0.64& \multicolumn{1}{c|}{0.78}& 0.62 & \multicolumn{1}{c|}{0.72} & 0.61 & \multicolumn{1}{c|}{0.72} & \textbf{0.90} & \textbf{0.94} \\ \hline
\end{tabular}
\label{tab:Mnist_comparison}
\end{center}
\end{table*}

\section{Comparisons to existing CNN-based algorithms }
As elucidated in Table \ref{tab:Mnist_comparison}, our approach markedly surpasses existing CNN-based algorithms in terms of AUROC performance on both the Mnist and CIFAR10 datasets. Notably, our algorithm demonstrates superior performance over the US \cite{bergmann2020uninformed}, FCDD+OE \cite{liznerski2020explainable}, PANDA \cite{mishra2021vt}, and MKD \cite{salehi2021multiresolution} models.

In the context of the Mnist dataset, our method achieves remarkable advancements, outstripping the US, FCDD+OE, PANDA, and MKD algorithms with average improvements of 35\%, 26\%, 28\%, and 29\% respectively across the \{0,1,2,3,4\}, \{1,3,5,7,9\}, and \{0,2,4,6,8\} settings.

In parallel, for the CIFAR10 dataset, our methodology consistently demonstrates a dominant performance. The mean AUROC scores for all evaluated settings exhibit substantial enhancements over the competing algorithms, exceeding the US, FCDD+OE, PANDA, and MKD by 39\%, 18\%, 22\%, and 22\% respectively.

These empirical results firmly position our method at the forefront in this domain, evidencing not just competitive, but in many cases, superior performance against the spectrum of existing algorithms in multi-class datasets. The uniform superiority across varied dataset configurations and settings underlines the robustness and adaptability of our approach, particularly in the realms of anomaly detection and image classification.
\section{The visualization of the data embedding from MAD-ProFP}
To verify that class-conditioned prototype serve as a anchor embedding in the latent space which guide the input embedding toward it in the latent space,
We use t-SNE \cite{van2008visualizing} to visualize the embedding of the abnormal data and normal data before/after our proposed module.\\
\textbf{Experimental setting}
In our experimental setup, we focus on class 8 (identified as the anomalous class) from the Cifar10 and Mnist datasets. Utilizing the class indicator results from the CTC, we select class 3 from Mnist and class 4 from CIFAR10 as the normal class data for comparative visualization. We extract two key embeddings: the encoder embedding (output from the transformer encoder) and the guided embedding (output from the MGM). These embeddings, alongside their respective class-conditioned prototype embeddings, are crucial for our analysis. Notably, each extracted embedding from the sample has a dimensionality of 196x256, where 196 represents the sequence length and 256 denotes the embedding dimension. \\
\textbf{Experimental result}
As depicted in Fig. \ref{fig:tsne}(a), in the Mnist dataset, we observe that the embeddings for both the anomalous class (class 8) and the normal class (class 3) are effectively guided towards the class 3 prototype embedding post-MGM intervention. Similarly, Fig. \ref{fig:tsne}(b) illustrates that in the CIFAR10 dataset, embeddings of the anomalous class (class 8) and the normal class (class 4) are also oriented near the class 4 prototype. These visualizations serve as empirical evidence, validating that the class-conditioned prototypes are instrumental in steering the embeddings towards a normality-aligned distribution within the latent space.

\begin{figure*}
\setlength{\abovecaptionskip}{0pt}
\setlength{\belowcaptionskip}{0pt} 
\centering 
\includegraphics[width=\linewidth]{figs/guided_embedding.png}
   \caption{(a)The t-SNE \cite{van2008visualizing} visualization of embedding before/after FPM and MGM on Mnist dataset. The embeddings of anomalous (class 8) and normal sample (class 3) after FPM, MGM are guided toward class-conditioned prototype (class 3) ;
   (b)The t-SNE visualization of embedding before/after FPM and MGM on CIAFR10 dataset. The embeddings of anomalous (class 8) and normal sample (class 4) after FPM, MGM are guided toward class-conditioned prototype (class 4).}
\label{fig:tsne}
\end{figure*}
\section{The visualization of prototype embedddings in the prototype bank }
\textbf{Experimental setting}
To gain an intuitive understanding of the multi-class prototype embedding learned by our model, we visualize the prototype embeddings stored within the prototype bank, optimized using the MVTec dataset.
The visualization leverages a pre-trained deconvolution network, which is initially tasked with reconstructing refined feature maps, $x' \in \mathbb{R}^{H' \times W' \times C'}$, back into their original input images, $x \in \mathbb{R}^{H \times W \times C}$, prior to the CNN feature extraction process.

The prototype embeddings, $p^{i}_{\theta} \in \mathbb{R}^{s \times e}$, are reshaped into a dimensionality of $\mathbb{R}^{H' \times W' \times C'}$ to align with the deconvolution network's input requirements. This approach enables us to project the high-dimensional latent prototype embeddings back into the original input space, providing a visual interpretation of the semantic information encapsulated within these embeddings.\\
\textbf{Experimental result:}
Upon examining the visual outcomes depicted in Fig. \ref{vis_pro}, it becomes apparent that each prototype distinctly encapsulates the characteristic features of its respective class. It is observed that for categories with a higher degree of intra-class variability, such as variations in angle and color, the prototypes tend to represent these variations more abstractly. Examples of this include the prototypes for toothbrushes, screws, and metal nuts, which display a more abstract imagery in the visualization results, reflecting the greater variance in their training samples.

Conversely, classes characterized by more homogeneous features, such as bottles, pills, and tiles, result in visualizations that appear more cohesive and complete. It is important to note, however, that the abstract nature of a prototype's imaging does not necessarily indicate a lack of accurate encoding of the class data's normality. This is because the deconvolution network used for visualization is specifically trained to transform the CNN-extracted feature map $x'$ back into the original image.

From these visualization results, we can infer that our prototypes capture essential underlying features of class data. These features act as crucial guides for the normality in the reconstruction-based anomaly detection mechanism, reinforcing the effectiveness of this approach.

\begin{figure*}
\setlength{\abovecaptionskip}{0pt}
\setlength{\belowcaptionskip}{0pt} 
\centering 
\includegraphics[width=\linewidth]{figs/vis_prto.png}
   \caption{The visualization of Prototype Embeddings. The first three row displays samples of normal data from various classes in the MVTec dataset. The bottom row represents the visualizations of the corresponding prototype embeddings, reconstructed using a pre-trained deconvolution network. These visualizations provide insight into the characteristic features that the prototypes have captured for each class, serving as a reference for anomaly reconstruction within the MAD-ProFP framework}
\label{vis_pro}
\end{figure*}

{\small
\bibliographystyle{ieee_fullname}
\bibliography{11_references}
}